\documentclass[12pt]{article}
\usepackage[utf8]{inputenc}
\usepackage{graphicx}
\usepackage{mathtools}
\usepackage{amsmath}
\usepackage{caption}
\usepackage{subcaption}
\usepackage{lipsum}
\usepackage{float}
\usepackage{booktabs, multirow}
\usepackage{bbold}
\usepackage{soul}

\usepackage[english]{babel}
\usepackage{authblk}
\usepackage{enumitem}
\usepackage{tikz}
\usetikzlibrary{tikzmark}

\usepackage{lipsum}

\usepackage[letterpaper,top=0.8in,bottom=0.8in,left=0.8in,right=0.8in]{geometry}

\usepackage[colorlinks=true, allcolors=blue]{hyperref}

\usepackage{rotating}

\title{Question-Answering System Extracts Information on Injection Drug Use from Clinical Notes}

\author[1,*]{Maria Mahbub}
\author[2]{Ian Goethert}
\author[3,4]{Ioana Danciu}
\author[2]{Kathryn Knight}
\author[1]{Sudarshan Srinivasan}
\author[5,6]{Suzanne Tamang}
\author[7]{Karine Rozenberg-Ben-Dror}
\author[5]{Hugo Solares}
\author[5]{Susana Martins}
\author[5]{Jodie Trafton}
\author[1]{Edmon Begoli}
\author[8]{Gregory D. Peterson}
\affil[1]{Cyber Resilience and Intelligence Division, Oak Ridge National Laboratory, Oak Ridge, TN, USA}
\affil[2]{Information Technology Services Division, Oak Ridge National Laboratory, Oak Ridge, TN, USA}
\affil[3]{Computational Sciences and Engineering Division, Oak Ridge National Laboratory, Oak Ridge, TN, USA}
\affil[4]{Department of Biomedical Informatics, Vanderbilt University, Nashville, TN, USA}
\affil[5]{Program Evaluation and Resource Center, Office of Mental Health and Suicide Prevention, Department of Veterans Affairs, USA}
\affil[6]{Department of Medicine, Stanford University School of Medicine, Stanford, CA, USA}
\affil[7]{Veterans Affairs Great Lakes Health Care System, Westchester, IL, USA}
\affil[8]{Department of Electrical Engineering and Computer Science, University of Tennessee, Knoxville, Knoxville, TN, USA}

\affil[*]{Corresponding author: mahbubm@ornl.gov}
\date{}

\begin{document}
\maketitle

\section*{Abstract}
\textbf{Background:} Injection drug use (IDU) is a dangerous health behavior that increases mortality and morbidity.
Identifying IDU early and initiating harm reduction interventions can benefit individuals at risk. 
However, extracting IDU behaviors from patients' electronic health records (EHR) is difficult because there is no International Classification of Disease (ICD) code and the only place IDU information can be indicated is unstructured free-text clinical notes.
Although natural language processing can efficiently extract this information from unstructured data, there are no validated tools.
\\
\textbf{Methods:} To address this gap in clinical information, we design and demonstrate a question-answering (QA) framework to extract information on IDU from clinical notes. 
Our framework involves two main steps: (1)~generating a gold-standard QA dataset and (2)~developing and testing the QA model.
We utilize 2323 clinical notes of 1145 patients sourced from the VA Corporate Data Warehouse to construct the gold-standard dataset for developing and evaluating the QA model.
We also demonstrate the QA model's ability to extract IDU-related information on temporally out-of-distribution data.
\\
\textbf{Results:} Here we show that for a strict match between gold-standard and predicted answers, the QA model achieves  51.65\% F1 score. For a relaxed match between the gold-standard and predicted answers, the QA model obtains 78.03\% F1 score, along with 85.38\% Precision and 79.02\% Recall scores.
Moreover, the QA model demonstrates consistent performance when subjected to temporally out-of-distribution data.
\\
\textbf{Conclusions:} Our study introduces a QA framework designed to extract IDU information from clinical notes, aiming to enhance the accurate and efficient detection of people who inject drugs, extract relevant information, and ultimately facilitate informed patient care.

\maketitle

\section*{Plain language summary}
Injection drug use (IDU) behavior poses significant health risks.
Therefore, it is crucial to identify IDU behavior early to offer timely assistance. 
However, extracting IDU information from patients' electronic health records (EHR) is challenging because the only place IDU information can be indicated is free-text clinical notes.
Manually extracting information from these notes is time-consuming and inefficient.
This study explores the ability of question-answering (QA) models in machine learning – whereby computer software is trained to read and understand these notes and extract IDU details.
Our findings illustrate that QA models achieve noteworthy performance in extracting information related to IDU from clinical notes.
Potentially, this approach can help healthcare providers efficiently and accurately identify people who inject drugs, extract information on IDU behavior, and provide better patient care.

\section*{Introduction}
\label{sec:intro}

Injection drug use (IDU) is a  critical health concern in the United States and internationally \cite{goel2016intravenous}.
Most people begin using illicit drugs through other modes of administration such as smoking, intranasal absorption, or oral ingestion.
As dependence grows, individuals tend to prefer the intravenous (IV) route of drug administration, injecting drugs directly into the veins, as it offers stronger and more immediate effects \cite{o2006drug}.
The number of people who inject drugs increased almost fivefold from 2011 to 2018 according to estimates in \cite{bradley2023estimated}, whereas the number of IDU-related overdoses increased eightfold from 2000 to 2018 \cite{hall2022estimated}.

IDU is a highly dangerous practice, which can lead to complicated medical conditions such as abscesses and cutaneous infections, scarring and needle tracks, endocarditis, HIV/AIDS, Hepatitis C, overdose and deaths \cite{cornford2016physical, goel2016intravenous, marks2022infectious, powell2019transitioning, strathdee2020preventing}.
An increase in IDU is also associated with an increase in morbidity and mortality \cite{wurcel2016increasing, sredl2020not, see2020national}.

Accurately identifying IDU behaviors in \textcolor{black}{people who inject drugs} is crucial for risk assessment and detection of patients that can benefit from harm reduction interventions to potentially prevent IDU-related morbidity and mortality \cite{goodman2022natural, edwards2014exploring}.
In the literature, the study of IDU-related information extraction has been performed along with other socio-behavioral determinants of health (SBDH). Considering and including SBDH such as prior incarceration, substance use (regardless of administration mode), treatment attitude, psychological distress, and interpersonal violence improve patient mortality and enhances the prediction of medication adherence, hospital readmission, and suicide attempts \cite{nijhawan2019clinical, chen2020social}.

Despite the growing interest, SBDH such as IDU is not identifiable in patients’ electronic health records (EHRs) through ICD codes; although not systematically assessed, it can be documented in clinical notes \cite{patra2021extracting, feller2020detecting}.
While structured data fields derived from EHRs may provide some amount of information about risky drug use behaviors and morbidities related to IDU, the clinical note is the only place it can be explicitly documented \cite{goodman2022natural}.
Despite being clinically meaningful and having the potential to identify patients that can benefit from harm reduction interventions, care providers often struggle to retrieve these data points from EHRs, and evidently, the exclusion of this data may result in an overall reduced quality of care \cite{gottlieb2015moving, weir2015qualitative}.

\begin{figure}[!htb]
    \centering
    \includegraphics[width = .8\textwidth]{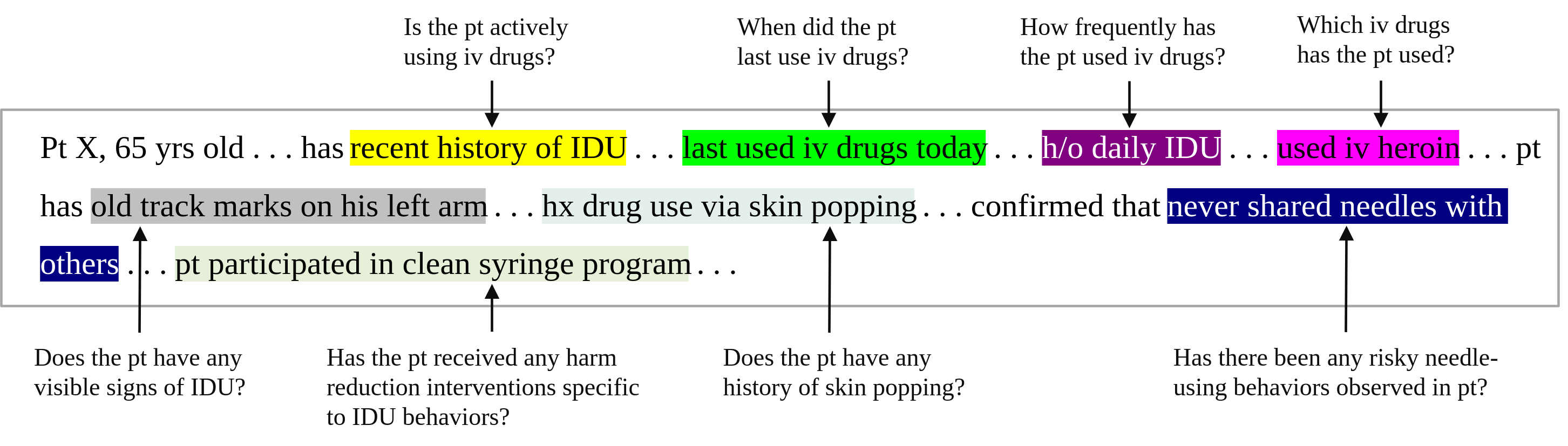}
    \caption{A sample clinical note featuring questions about the  IDU behavior \textcolor{black}{in people who inject drugs,} with extracted IDU-related information color-coded in the note.}
    \label{fig:example}
\end{figure}

Natural language processing (NLP) can help extract SBDH-related information from clinical notes and expand the utility of such information in patient care \cite{hayes2022using, topaz2019extracting, peng2023clinical}.
NLP is a branch of computer science that involves automated learning, understanding, and generation of natural languages, enabling the interactions between machines and human languages. 
Although NLP deals with a variety of tasks involving unstructured text data (e.g., event prediction \cite{mahbub2022unstructured}, entity recognition \cite{li2020survey}, question-answering (QA) for information extraction \cite{mahbub2023cpgqa}, and relation extraction \cite{eberts2019span}), in this article, we use extractive question-answering (extractive QA) task to automatically extract information related to IDU from clinical notes in EHRs. To avoid redundancy, for the rest of the paper we use ``QA'' in place of ``extractive QA''.
In this QA task, given a query and a clinical note, a QA model would return the relevant answer verbatim from the note as the extracted information.
Thus, a QA system is tasked with learning to read and comprehend the clinical note provided a query and then extract information consisting of consecutive words from the notes relevant to the query from that note (Figure~\ref{fig:example}).

\textcolor{black}{We use QA to address the information extraction problem for the following reasons. The texts in the clinical notes are very unstructured in nature, for example, the information regarding injection drug names can be presented in the notes in multiple forms, such as “opioids: denies recent use, hx ivdu\footnote{ivdu: intravenous drug use}, claims last use years ago. other drugs: hx methamphetamine use, has been using daily via injecting since relapse in December”, “ivdu (cocaine/methamphetamine)”, ”reports using iv meth”, “iv cocaine mixed with heroine use”, “used meth by iv drug use”,  or “history of daily heroine use, prior ivdu”. Given the demonstrated success of QA models in extracting information of diverse forms from clinical notes \cite{pampari2018emrqa}, we chose to focus on the QA task in NLP. Moreover, one potential implementation of this work would be to incorporate the developed model into a chatbot framework, enabling clinicians to inquire about IDU behavior in people who inject drugs at the point of care by posing questions with various syntactic structures. It would help clinicians identify \textcolor{black}{people who inject drugs} and pinpoint related status.}

Although not specific to IDU, several studies have focused on identifying clinical concepts or information on substance use disorders (SUD) using NLP \cite{wang2015automated, peng2023clinical}.
In these studies, various NLP techniques have been used to extract SUD-related information.
The stemming algorithm has been used to identify words and phrases associated with mental illness and substance use in clinical notes \cite{ridgway2021natural, nadkarni2011natural}. 
Dependency structure has been utilized to capture relationships between phrases and tokens in the substance use statement \cite{wang2015automated}.
Word-embedding models have been employed to identify alcohol and substance abuse status \cite{topaz2019extracting}.
Machine reading comprehension has been applied to extract some clinical concept categories and relation categories, such as relations of medications with adverse drug events and SBDH \cite{peng2023clinical}.
Multi-label text classification and sequence labeling have been used to identify sentences containing labeled arguments about drug use \cite{torii2023task}.
Topic modeling and keyword matching techniques have been leveraged to extract drug use--related information \cite{feller2018using}.
Techniques such as active learning \cite{lybarger2021annotating}, multi-label classification \cite{han2022classifying, yu2022assessing, feller2018towards, ahsan2021mimic}, concept extraction, and joint extraction of entities and relations have been employed to extract information about drug use \cite{lybarger2022leveraging}.
Researchers have also focused on identifying drug use information by using NLP-specific techniques to detect opioid use disorder and predict overdose \cite{carrell2015using, afshar2021external, afshar2022development, lingeman2017detecting, blackley2020using, zhu2022automatically, poulsen2022classifying, ward2019enhancing, badger2019machine, hazlehurst2019using, harris2020challenges, goodman2022development}.
In the literature, we came across one research study that has focused exclusively on IDU. The study has utilized rule-based algorithms, such as regular expressions (RegEx), NegEx \cite{chapman2001simple}, and N-grams to search for \textcolor{black}{very limited} IDU-related terms, with the objective of identifying people who inject drugs (PWIDs) \cite{goodman2022natural}.
\textcolor{black}{In our study, on the other hand, we focus on extracting a broad spectrum of information on injection drug use from clinical notes. This encompasses details such as drug names, active/historical use, frequency of use, risky needle-using behavior, visible signs of IDU, last use, skin popping, harm reduction interventions, and existence of IDU.}
Since evidence of IDU cannot be found in structured EHR data and therefore must be inferred from clinical notes, this study's sole focus on IDU aims to help understand how this phenomenon is represented in unstructured notes data and augment techniques that have used NLP techniques less generalizable to this population.
To the best of our knowledge, to date, there has been no published attempt at developing a QA algorithm to extract IDU-related information from clinical notes.

\textcolor{black}{To solve the QA task, we use transformer-based deep learning models \cite{lee2020biobert, devlin2018bert} that are known to be one of the most streamlined ways to solve QA tasks and achieve comparable performance in extracting targeted information from different types of biomedical documents, such as scholarly articles \cite{lee2020biobert, mahbub2022bioadapt}, clinical practice guidelines \cite{mahbub2023cpgqa}, electronic medical records \cite{pampari2018emrqa}, etc.
Nonetheless, evidence suggests that supervised deep learning models require high-quality and large-scale annotated datasets to achieve good performance in any task \cite{rajpurkar-etal-2016-squad, joshi-etal-2017-triviaqa, devlin2018bert} and the absence of such a dataset for our targeted QA task poses a critical challenge.
An annotated QA dataset comprises data samples, with each sample containing a context (e.g., a clinical note), a question, and an answer extracted verbatim from the context (i.e., the extracted information).
In addressing the challenge posed by the limited availability of annotated QA data for constructing an effective QA model, our study takes a two-fold approach.
First, we built a high-quality gold-standard QA dataset in collaboration with a subject matter expert (SME), facilitating model training and testing.
The dataset includes clinical notes as contexts and question-answer pairs specific to IDU.
Then, using this meticulously curated gold-standard dataset, we dive into the primary objective of this study -- develop and assess the QA system for IDU-related information extraction from clinical notes.
We also perform an error analysis to identify the strengths and weaknesses of our QA system, providing valuable insights to guide future research endeavors.
}

\section*{Methods}
\begin{figure}[htbp]
    \centering
    \includegraphics[width = .8\textwidth]{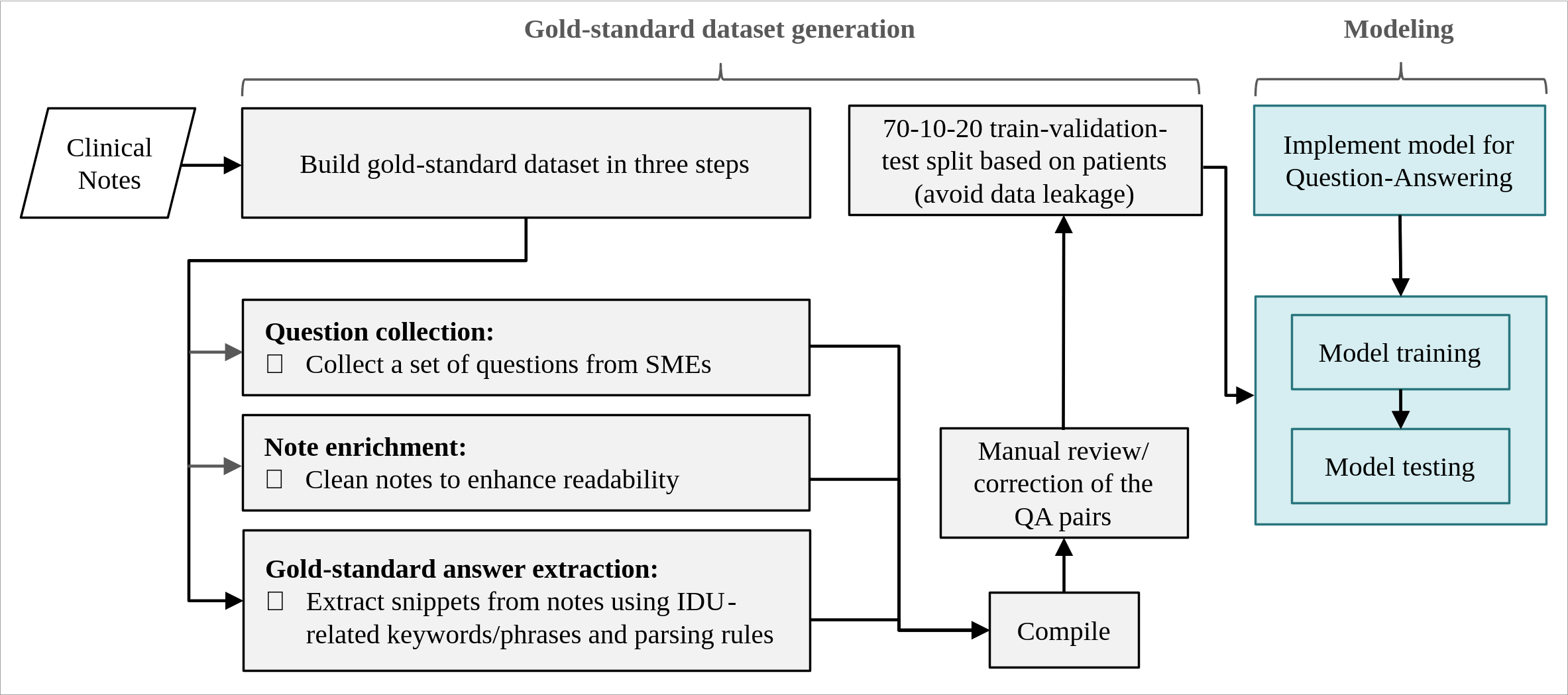}
    \caption{\textcolor{black}{Framework for a two-part study for extracting information on IDU behavior in \textcolor{black}{people who inject drugs} from clinical notes. The first part consists of gold-standard dataset generation in three primary steps and the second part consists of QA model development from implementation to inference.}}
    \label{fig:framework}
\end{figure}

In this section, we elaborate on the formulation of this study and its two components: (i) Gold-standard dataset generation and (ii) modeling  (Figure~\ref{fig:framework}).
\textcolor{black}{Furthermore, we outline the specifications of the gold-standard dataset, the experimental setup, and the metrics used to assess the performance of the QA models.}

\subsection*{Problem formulation}
We formulate the information extraction task as a QA problem in NLP in the following manner:
Given a question on patients' behavior about IDU and a clinical note with IDU-related information (i.e., the context), a QA system retrieves the relevant information (i.e., the answer) from the provided note.

For example, given the question---\textcolor{black}{\textit{Does the patient have a history of IDU?}} and the clinical note---\textcolor{black}{\textit{pt X, 200 yrs old .~.~. he has a history of smoking with 50 pack years, quit 10~years ago .~.~. social ethanol user .~.~. no history of idu .~.~. remote history of marijuana use .~.~. family hx: .~.~. physical exam: .~.~. provider: name.}}---the QA system is expected to return the answer---\textcolor{black}{\textit{no history of idu}}---verbatim from the note.

\subsection*{Gold-standard dataset generation}
\label{sec:method_dataset}

QA is a supervised NLP learning task and as such requires an annotated gold-standard dataset for model development and inference.
In a QA dataset, each sample \textcolor{black}{consists}  of the context, a question, and an answer, with the question-answer pairs serving as annotations.
To generate a gold-standard dataset from clinical notes, which serve as the context, we employ a three-stage process outlined in Figure~\ref{fig:framework}: (1) question collection, (2) \textcolor{black}{note enrichment}, and (3) gold-standard answer extraction.

\subsubsection*{Question collection}
\label{sec:method_question}
We initialize the process of question collection for the dataset by asking SMEs about the kind of information on IDU they are interested in from the clinical notes.
We then generate a set of questions based on their interest.
Table~\ref{tab:question} shows the nine categories of interest.
In the rest of the paper, we use the term ``Query Group'' to imply categories of interest.
Table~\ref{tab:question} also provides sample questions and answers for each query group.

\begin{table}[!htb]
\centering
\begin{tabular}{p{0.29\linewidth}  p{.66\linewidth}}\hline
\textbf{Query Groups}                        & \textbf{Sample Question} \newline \textbf{$\rightarrow$ Sample Answer} \\\hline
Drug names                          & Which iv drugs has the patient used? \newline                       
                                      $\rightarrow$ hx of iv heroin abuse, cocaine, and bnz \\\hline
Visible signs of IDU          & Does the patient have any needle track marks? \newline                       
                                      $\rightarrow$ track marks noted over bilateral upper extremities \\\hline
Risky needle-using behavior         & Has the patient ever shared needles? \newline                       
                                      $\rightarrow$ hx of ivdu, has shared needles in the past few weeks \\\hline
Active/historical use            & Is the patient actively using iv drugs? \newline                       
                                      $\rightarrow$ h/o active iv drug use \\\hline
Frequency of use                    & How frequently has the patient used iv drug? \newline                       
                                      $\rightarrow$ history of ivdu (reports daily use of heroin) \\\hline
Last use                            & When did the patient last use iv drugs? \newline                       
                                      $\rightarrow$ daily use of iv heroin with last use 4 days prior to admission \\\hline
Skin popping                        & Does the patient have any history of skin popping? \newline                       
                                      $\rightarrow$ diffuse scarring from skin popping on lower extremities \\\hline
Harm reduction \newline
interventions                       & Has the patient been counseled on safe injection technique? \newline                       
                                      $\rightarrow$ discussed the importance of using clean needles with patient should he   continue to inject drugs \\\hline
Existence of IDU                   & Does the patient have any history of IDU? \newline                       
                                      $\rightarrow$ no ivdu or h/o sharing needles\\\hline                                                                   
\end{tabular}
\caption{Types of information about IDU that are most likely to be inquired by clinicians from the clinical notes, categorized into nine query groups. Abbreviations: ivdu, intravenous drug use; bnz, benzodiazepines.}
\label{tab:question}
\end{table}

Each query group targets to extract one category of information from the notes pertaining to that group.
For example, the query group ``drug names'' targets to extract any information about IV drug names from the notes.
\textcolor{black}{In our gold-standard dataset, we include multiple variations of questions for each query group.
For example, for the query group ``drug names'', we have five different variations of questions as follows: ``To what IV drugs has the patient been exposed?'', ``Which IV drugs has the pt used?'', ``Which intravenous drugs has the patient used?'', ``Which injection drugs?'', ``Which illicit drugs has the patient injected?''.}

\textcolor{black}{We do this for the following reasons. We anticipate our system to be used as a standalone application -- a more user-friendly QA tool to collect IDU evidence -- and to be capable of handling different variations of questions posed by clinicians.
Furthermore, we hope that different variations of questions for each query group will help increase the QA model's user-flexibility, comprehensiveness, and robustness, ultimately enhancing its performance in real-world applications, as follows:
(i) Users may pose questions in different ways based on their preferences or understanding. A QA model trained with diverse question variations is more adaptable and capable of accommodating the linguistic diversity inherent in user queries.
(ii) Including variations of questions during training helps the QA model become more robust by exposing it to diverse ways the same question can be asked, preparing the model to handle real-world scenarios where questions may be phrased differently but still seek the same information.
(iii) Variations of questions during training enable the QA model to generalize its understanding. Instead of memorizing specific phrasings, the model learns the underlying patterns and associations between questions and answers, improving its ability to respond accurately to novel queries.
}

We use abbreviations, synonyms, and syntactical variations to introduce variations in the questions for each query group, as follows:
\paragraph{Abbreviations:}
``Is the \underline{patient} actively using intravenous drugs?'' $\rightarrow$ ``Is the \underline{pt} actively using intravenous drugs?'',
``Is the patient actively using \underline{intravenous} drugs?'' $\rightarrow$ ``Is the patient actively using \underline{iv} drugs?'', etc.

\paragraph{Synonyms:}
``Does the pt have a history of using \underline{intravenous} drugs?'' $\rightarrow$ ``Does the pt have a history of using \underline{injection} drugs?'',
``Does the pt have a history of \underline{IDU}?'' $\rightarrow$ ``Does the pt have a history of \underline{IVDU}?'',

\paragraph{Syntactical variations:}
``Which iv drugs has the patient used?'' $\rightarrow$ ``To which iv drugs has the patient been exposed?'',
``Does the pt have a history of IVDU?'' $\rightarrow$ ``Has the pt ever used IV drugs?''

It should be noted that when identifying abbreviations and synonyms to be used in questions, we only choose terms and variants that clinicians commonly use. Examples of these terms and variants include ``patient'' and ``pt'', ``intravenous'' and ``iv'', ``history'' and ``hx'', and ``IVDU'' and ``IDU''.
And, to ensure that we were able to accurately capture the nuances of possible language usage in the questions with regard to syntactical variations, we sought the guidance of SMEs.

\subsubsection*{\textcolor{black}{Note enrichment}}
\label{sec:method_context}
The contexts in the gold-standard dataset are clinical notes that contain some IDU-related information.
As such, we select a cohort of patients whose notes have a higher chance of containing IDU-related information, such as patients who have been diagnosed with Hepatitis C.
To guarantee that the clinical notes include information relevant to IDU and narrow down the notes accordingly, we use a list of keywords/phrases that are indicative of IDU (refer to Table~\ref{tab:keywords}) and has been developed by SMEs.
SMEs followed an iterative approach to create this list. 
They began by compiling a list of common terms related to IDU, which they then refined by reviewing the associated snippet.
They removed terms that caused excessive noise, such as ``slamming'' and ``drug paraphernalia,'' and added terms like ``skin popping'' to enhance granularity.
The experts received extensive training to sort and/or define the snippet categories, and they validated the terms to ensure their accuracy.

For our study, we assumed that the presence of any of these IDU-related keywords indicates the presence of relevant information pertaining to IDU in the note.
Hence, we discard the notes that do not contain any of the words/phrases provided in Table~\ref{tab:keywords} suggesting the possible non-existence of any IDU-related information in that note.
As shown in Table~\ref{tab:keywords}, this list can be categorized into the following groups: IV drug names, visible signs of IDU, risky needle-using behavior, skin popping, harm reduction interventions, and generic IDU terms.

\begin{table}[htb]
    \centering
    \begin{tabular}{p{0.3\linewidth}  p{0.65\linewidth}}\hline
      \textbf{Keyword Groups}  & \textbf{Keywords/phrases} \\ \hline
      IV drug names & iv/intravenous/inject(s/ed) heroin/meth/cocaine/crack, speedball \\\hline
      Visible signs of IDU & track marks, skin popping \\\hline
      Risky needle-using behavior & sharing/shared/dirty needle \\\hline
      Skin popping & skin popping \\\hline
      Harm reduction\newline interventions & community/clean/safe syringe service/program, ssp, ris4e, counseled on safe(r) injection, safe injection technique \\\hline
      Generic IDU terms & ivdu, idu, ivda, iv/intravenous/injection drug use/abuse, inject/injected drug, drug(s) by injection, iv/intravenous drug injector/injection, illicit iv/intravenous drug, iv/intravenous drug paraphernalia, suspect injecting, pwid \\\hline
    \end{tabular}
    \caption{A list of IDU keywords/phrases provided by SMEs.
    Abbreviations: ssp, syringe services programs; ivda, intravenous drug abuse; ris4e, resists infection by sterile syringe safe sex and education; PWID, people who inject drugs.}
    \label{tab:keywords}
\end{table}

\textcolor{black}{To enhance the readability of clinical notes and make them more suitable for automated processing, we conduct rigorous manual exploration of the final set of notes, identifying some common patterns that can help clean them using RegEx.
It is important to note that to preserve crucial information in the clinical notes, we perform minimal data cleaning, as follows:
(i) Remove newlines following within-sentence punctuation marks, such as comma, semicolon, or colon. For instance, removing the newline (``\textbackslash n'') highlighted in the sentence ``Veteran reported using iv meth,\textbf{\textbackslash n} iv cocaine and etoh.''.
(ii) Remove newlines appearing before punctuation marks, such as period, comma, or semicolon. For example, removing the newline (``\textbackslash n'') highlighted in the sentence ``Veteran reported using iv meth, iv cocaine and etoh\textbf{\textbackslash n}.''.
(iii) Remove newlines positioned between words within the same sentence. For example, removing the newline (``\textbackslash n'') highlighted in the sentence ``Veteran reported\textbf{\textbackslash n} using iv meth, iv cocaine and etoh.''.
(iv) Consolidate multiple consecutive occurrences of newlines, white spaces, or punctuations into single instances. For example, replacing multiple periods with a single period in ``Veteran reported using iv meth, iv cocaine and etoh\textbf{.............}''.
We perform these steps to clean all the notes used for training, validation, and testing.
}

\subsubsection*{Gold-standard answer extraction}
\label{sec:method_answer}

The next step in our dataset generation process is to extract gold-standard answers (i.e., information related to IDU) from the clinical notes.
Clinical notes are inherently lengthy, and manually extracting the gold-standard answers from them requires a substantial amount of time, rendering the process unfeasible.
Therefore, we devise a pre-annotation strategy involving an automated step-by-step answer extraction process that integrates rule-based NLP techniques.
The primary objective of this phase is to substantially reduce the manual annotation/review effort.
Nevertheless, to ensure the utmost quality of the gold-standard dataset, the outputs from this pre-annotation phase, along with the associated questions, underwent subsequent manual review and correction by a subject-matter expert with a PhD in Psychology and an extensive background in substance use disorder, counseling, and treatment.
Our pre-annotation strategy is based on three assumptions:
\paragraph{Assumption 1:} Our QA task only tackles information extraction (i.e., answering questions) from one single place (a sentence) in the note at a time.

\paragraph{Assumption 2:} The inquired information can be found in a single sentence in the note.
This assumption stems from our rigorous manual exploration of the notes during the note enrichment step, where we find RegEx patterns.
Our observation indicates that, in most instances, a single sentence per question suffices to capture the relevant answer. Nonetheless, we acknowledge that this straightforward sentence selection process may not always be optimal. Unstructured clinical notes often deviate from grammatical rules. Additionally, information presentation in these notes may vary, adopting styles such as questionnaires or bulleted lists.
As a result, a single sentence in the traditional sense occasionally leads to either a larger text segment or a fragmented part of a single piece of information. These instances lead to the inclusion of irrelevant or incomplete information in the answers, and we address and rectify these issues during our manual review phase.

\paragraph{Assumption 3:} If the note contains IDU-related information in multiple locations, each is considered a separate answer string.
Furthermore, multiple answer strings from the same note are expected to contain different kinds of information that should be answered by different questions.
For example, in the note snippet---\textcolor{black}{\textit{pt has a history of smoking with 50 pack years, quit 10~years ago .~.~. social ethanol user .~.~. has h/o ivdu .~.~. remote history of marijuana use .~.~. last used iv meth 2~years ago .~.~.}}, there are two locations where IDU-related information can be found -- \textcolor{black}{\textit{has h/o ivdu}} and \textcolor{black}{\textit{last used iv drugs 2~years ago}}.
In such cases, we consider them as separate answers that are retrieved when asked the following questions: \textcolor{black}{\textit{Does the pt have a history of IDU?}} and \textcolor{black}{\textit{When did the pt last use IV drugs?}}.

Given clinical notes, we extract the automated gold-standard answers using rule-based NLP techniques as follows:

\paragraph{Step 1:} Tokenize the sentences in the notes.
Here, we define ``sentence'' in the traditional sense, ending with a period.
Therefore, for the sentence tokenization, we use periods to indicate the end-of-sentence.

\paragraph{Step 2:} Identify sentences that contain any of the IDU keywords from Table~\ref{tab:keywords} using regular expression string matching and discard the rest.

\paragraph{Step 3:} At this point, the sentences containing the IDU keywords can be ideally considered the gold-standard answers (i.e., extracted information relevant to IDU). 
Nonetheless, our primary aim is to extract IDU-related information from the notes, but we also want the extracted information to be as precise as possible containing lesser nonessential information.
A full-sentence answer is most likely to include \textcolor{black}{nonessential information}, which can be further reduced by using parsing rules. Parsing rules refer to NLP techniques that can identify specific patterns of text within a string that represent the concepts of interest, while ignoring the remaining text.
An example of removing nonessential information from the answer can be transforming the sentence \textcolor{black}{\textit{social history: pt lives with family in [location], quit smoking 10~y ago, occ etoh, .... hx methamphetamine use, has been using daily via injecting since relapse in December.}} into the phrase \textcolor{black}{\textit{hx methamphetamine use, has been using daily via injecting since relapse in December}}.

To create the parsing rules in this study, we randomly sample a set of sentences and focus on identifying specific phrases that occur together before or after the IDU keywords and modify or provide information that is crucial to the IDU-related history of the patient (refer to Table~\ref{tab:question}).
These phrases can be adjacent to or distant from the keywords.
For example, \textcolor{black}{\textit{pt. lives with family, \underline{denies ivdu}.}} vs \textcolor{black}{\textit{pt lives with family, \underline{denies any tobacco, etoh or ivdu}.}}
In this example, the phrase ``denies'' provides crucial information on the IDU behavior of the patient.

In Table~\ref{tab:rule}, we provide a detailed list of these phrases along with the targeted pattern type, parsing rules, and examples of how they help reduce the \textcolor{black}{nonessential information} from the answers.
The parsing rules mainly focus on identifying patterns stating negative IDU mentions, temporal information, opioid use disorder specific to IDU, and status of track marks.

\textcolor{black}{Although these parsing rules can extract the correct concise gold-standard answers from the clinical notes in numerous cases, manual review reveals instances where the rules failed to accurately identify these answers.
This discrepancy was primarily attributed to the unstructured nature of information within the notes.}

\begin{sidewaystable}[!htbp]
\centering
\begin{tabular}{p{0.2\linewidth}  p{0.3\linewidth}  p{0.27\linewidth}  p{0.23\linewidth}}\hline
\textbf{Targeted Patterns}  & \textbf{Co-occurring Phrases\newline Before/After IDU Keywords}  & \textbf{Parsing Rules}  & \textbf{Example Answers Before and After Parsing}
\\ \hline

Negations \newline (nullifying the existence of IDU/skin popping/track marks/needle sharing)
& Negation Phrases (NP): \newline den(ying/ies/ied), no, never
& \textbf{Sentence:} $<$text before NP$>$ \underline{NP} $<$text after NP containing keywords$>$ \newline
\textbf{Rule:} \underline{NP} $<$text after NP containing keywords$>$
& \textbf{Before:} 65y/o m w cardiac procedures, or recent surgical procedures, admits to drinking alcohol daily for the past 10 years, \underline{denies} any history of ivdu \newline
\textbf{After:} \underline{denies} any history of ivdu
\\ \hline

Temporal information \newline (active/historical/last/ frequency of use)
& Temporal Phrases (TemP): \newline past/remote/distant/prior/previous/ former/active/current/recent/last/long history/hx (or h/o), daily, occasional, past, remote, distant, prior, previous, former, active, current, recent, last, long, intermittent, hpi, history, hx, h/o, pmh
& \textbf{Sentence:} $<$text before TemP$>$ \underline{TemP} $<$text after TemP containing keywords$>$ \newline
\textbf{Rule:} \underline{TemP} $<$text after TemP containing keywords$>$
& \textbf{Before:} pt smokes cannabis, has a \underline{h/o} ivdu but none now, went to rehab 2070 \newline
\textbf{After:} \underline{h/o} ivdu but none now, went to rehab 2070
\\ \hline

Additional temporal \newline information
& Additional Temporal Phrases (ATP): \newline year(s)/yr(s)/month(s)/mnth(s)/ day(s)/d/wk/wks/mos ago
& \textbf{Sentence:} $<$TemP$>$ $<$text after TemP and before ATP containing keywords$>$ \underline{ATP} $<$text after ATP$>$ \newline
\textbf{Rule:} $<$TemP$>$ $<$text after TemP and before ATP containing keywords$>$ \underline{ATP}
& \textbf{Before:} last ivdu was 10 \underline{days ago}, snorts cocaine occasionally \newline
\textbf{After:} last ivdu was 10 \underline{days ago} 
\\ \hline

Opioid/substance use \newline disorder specific to IDU
& Phrases related to Substance use \newline disorder (SP): \newline substance/polysubstance use/abuse disorder, sud, psud, oud, polysubstance, opioid use disorder, opioid, opiate
& \textbf{Sentence:} $<$text before SP$>$ \underline{SP} $<$text after SP containing keywords$>$ \newline
\textbf{Rule:} \underline{SP} $<$text after SP containing keywords$>$
& \textbf{Before:} 200m w niddm, htn, bipolar disorder and \underline{oud} (iv heroin) on methadone maintenance, recent heroin relapse \newline
\textbf{After:} \underline{oud} (iv heroin) on methadone maintenance, recent heroin relapse
\\ \hline

Status of track marks
& Phrases related to Track Mark status (TMP): \newline arm(s)/abnormal/multiple/many/ \newline several/healing/healed/old/diffuse/ \newline localized/visible/red/iv/fresh/dark/ \newline needle/notable
& \textbf{Sentence:} $<$text before TMP$>$ \underline{TMP} $<$text after TMP containing ``track marks''$>$ \newline
\textbf{Rule:} \underline{TMP} $<$text after TMP containing ``track marks''$>$
& \textbf{Before:} comments: extremities: mid line in upper right arm, scars and \underline{old} track marks noted on mid arm \newline
\textbf{After:} \underline{old} track marks noted on mid arm
\\ \hline

\end{tabular}
\caption{A detailed list of co-occurring phrases before/after IDU keywords along with the parsing rules, targeted pattern type, and examples of answers before and after parsing}
\label{tab:rule}
\end{sidewaystable}

\subsubsection*{Question-to-answer mapping}

Finally, to generate the labels (question-answer pairs) of our gold-standard dataset, we create mappings between the questions from Section~\hyperref[sec:method_question]{Question collection} and the gold-standard answers from Section~\hyperref[sec:method_answer]{Gold-standard answer extraction}.
We achieve this by considering the query groups in Table~\ref{tab:question}.
For each query group, we identify a group of words in the gold-standard answers that are most likely to provide the information inquired by that query group.
\textcolor{black}{To compile this group of words, we engage in meticulous manual exploration, reading sentences containing IDU-keywords.}
Depending on the kind of information we are interested in (reflected by the query groups), these words can be either the keywords in Table~\ref{tab:keywords} or the words (Table~\ref{tab:rule}) that co-occur with the keywords and can help convey the information inquired by the user.
For example, co-occurring words ``daily'' and ``last'' describes the ``frequency of use'' and the ``last use'' of IDU, respectively.

Thus, for each query group, we decide on a group of words that are most likely to help convey the inquired information and map the answers that contain these words to the questions in that query group (Table~\ref{tab:question}).
\textcolor{black}{The resulting compilation is presented in the ``Words in Gold-standard Answers Most Likely to Provide Inquired Information'' column of Table \ref{tab:qa_map}. It is important to note, however, that this list is not exhaustive and represents only what we observe during our exploration, not an all-encompassing collection of potential phrases indicating the inquired information. Considering this, in our manual review phase, we manually correct annotations that are overlooked or mislabeled by these rules.
}

In Table~\ref{tab:qa_map}, we present the mappings between the query groups and the words in gold-standard answers that are most likely to provide inquired information.
We also demonstrate sample answers for each mapping.
Note that, the answers in one query group and the answers in a different query may not be mutually exclusive.
This is because, if we find words in an answer that belong to multiple query groups, then that answer is mapped to all questions from these query groups.
For example, the first sample answer from Table~\ref{tab:qa_map}---``recent ivdu with meth and heroin''---contains the words ``recent'' and ``heroin''/``meth'' from query groups ``active/historical use'' and ``drug names'', respectively.
Hence, this answer will be mapped to all questions in these two query groups.

\textcolor{black}{The well-known ConText rules \cite{harkema2009context} in the literature use a similar rule-based approach to identify the negation or temporality of a condition. They used a specific set of words tailored to the types of notes used in their study.
On the contrary, although the words utilized in our study share some commonalities, they exhibit notable differences from those employed in the ConText algorithm.
This distinction arises from variations in the notes used in our experiments and the specific information we target to extract from the notes.
Our study exclusively focuses on injection drug use. In contrast, the error analysis of ConText indicates its unsatisfactory performance in identifying temporality related to ``chronic conditions and risk factors, i.e., alcohol, drug'' in clinical notes. Additionally, while ConText explicitly identifies historical versus recent conditions, our question-answering system concentrates on extracting any temporal information regarding injection drug use, leaving the determination of whether the status is recent or historical to clinicians.}

\textcolor{black}{Regarding the query group ``last use'', it is crucial to note that a patient may have multiple note entries, each with its own last use. Given our study's emphasis on extracting information from one clinical note at a time, the definition of ``last use'' is confined to ``last use per note''.}

\begin{table}[!htb]
\centering
\begin{tabular}{p{0.24\linewidth} p{.4\linewidth} p{0.36\linewidth}}\\\hline
\textbf{Query Groups}
& \textbf{Words in Gold-standard Answers Most Likely to Provide Inquired Information}
& \textbf{Sample Answers}                                    
\\\hline
Drug names                        & IV drug names from Table~\ref{tab:keywords}, opioid, opiate, oud                        & recent ivdu with \underline{meth} and \underline{heroin}                     \\\hline
Visible signs of IDU             & Phrases for visible signs of IDU from Table~\ref{tab:keywords}                      & multiple \underline{track marks} over extremities                    \\\hline
Risky needle-using\newline behavior & Phrases for risky needle-using behavior from Table~\ref{tab:keywords}              & h/o \underline{sharing needles} with gf                              \\\hline
Active/historical use          & Temporal phrases from Table~\ref{tab:rule}, remission                                           & \underline{active} iv drug user up to day of admission               \\\hline
Frequency of use                  & daily, occasional, regularly, often, sometimes, frequently, intermittent                           & - iv cocaine \underline{daily}, $\sim$\$5--40/day                     \\\hline
Last use                          & Additional temporal phrases from Table~\ref{tab:rule}, last, quit, since, clean      & \underline{last} ivdu $>$30 years ago                                \\\hline
Skin popping                      & Phrases for skin popping from Table~\ref{tab:keywords}                               & h/o drug injections - \underline{skin popping}                       \\\hline
Harm reduction\newline interventions & Phrases for harm reduction interventions from Table~\ref{tab:keywords}  & patient participates in \underline{clean syringe program} \\\hline
Existence of IDU                 & Negation phrases from Table~\ref{tab:rule}                                           & \underline{denies} any ivdu for many years                          \\\hline
Existence of IDU                 & Remaining answers                 & iv drug user                          \\\hline

\end{tabular}
\caption{Mappings between the query groups and the words in gold-standard answers most likely to provide inquired information and example answers for each mapping}
\label{tab:qa_map}
\end{table}

\textcolor{black}{After generating the labels (i.e., question-answer pairs), we manually review the whole dataset in collaboration with a subject-matter expert to ensure that our gold-standard dataset is of high quality and accuracy.}

\subsection*{Modeling with question-answering system}
\label{sec:modeling}

In the next step of our study, we develop the QA model for extracting IDU-related information using the gold-standard QA dataset from Section~\hyperref[sec:method_dataset]{Gold-standard dataset generation}.
We use Bidirectional Encoder Representations from Transformers (BERT) \cite{devlin2018bert}-based deep learning QA models where the feature extractor is a trainable pre-trained BERT-based language model and the QA task layer is a single-layer feed-forward neural network.

We experiment with four state-of-the-art pre-trained language models---BERT \cite{devlin2018bert}, BioBERT \cite{lee2020biobert}, BlueBERT \cite{peng2019transferbluebert}, and ClinicalBERT \cite{alsentzer2019publiclyclinicalbert}---as trainable feature extractors and develop four QA models.

Provided a sequence of tokens (words or pieces of words) in a question and a clinical note, the QA model returns the start and end token of the answer span.
Any text between the start and end tokens included is then considered as the answer (i.e., the extracted information).
Together with the question and the note, the maximum allowable number of input tokens in these BERT-based QA models is 512.
To handle samples with longer clinical notes, we follow a widely known technique in QA modeling---sliding window with a document stride \cite{devlin2018bert}.

Below we provide a brief description of this technique:
Given an input question consisting of 20 tokens, the remaining allowable number of input tokens for the note is limited to 492 (which is 512 minus the 20 tokens in the question). If the note exceeds this limit, we employ a sliding window technique to split it into multiple chunks using a document stride of 128 tokens. The document stride determines the starting token of each subsequent chunk.
After this preprocessing step, each chunk prepended with the original question tokens is considered a separate data sample.

\subsection*{Dataset statistics}

We use clinical notes sourced from the VA Corporate Data Warehouse (CDW) to construct the gold-standard dataset.
\textcolor{black}{The selected notes correspond to the period of January 2022 and belong to patients with the Hepatitis C diagnosis.}
The identification of Hepatitis C positive patients is performed using ICD-10 codes.
\textcolor{black}{We select the cohort of patients with Hepatitis C as their clinical notes are more likely to include information related to IDU.
As explained in Section \hyperref[sec:method_context]{Note enrichment}, we narrow down the clinical notes using a list of keywords/phrases indicative of IDU (refer to Table~\ref{tab:keywords}).
}

To reduce computational overhead during training and because unusually large notes (determined by the outliers in the distribution of note lengths) may contain templated nonessential information that is not relevant to any specific patient, we remove some outlier notes based on the interquartile range of the note lengths.
We later show in Section~\hyperref[sec:error_analysis]{Error analysis} that note length does not affect the performance of the model at the time of inference.

\begin{figure}[htbp]
    \centering
    \includegraphics[width = \textwidth]{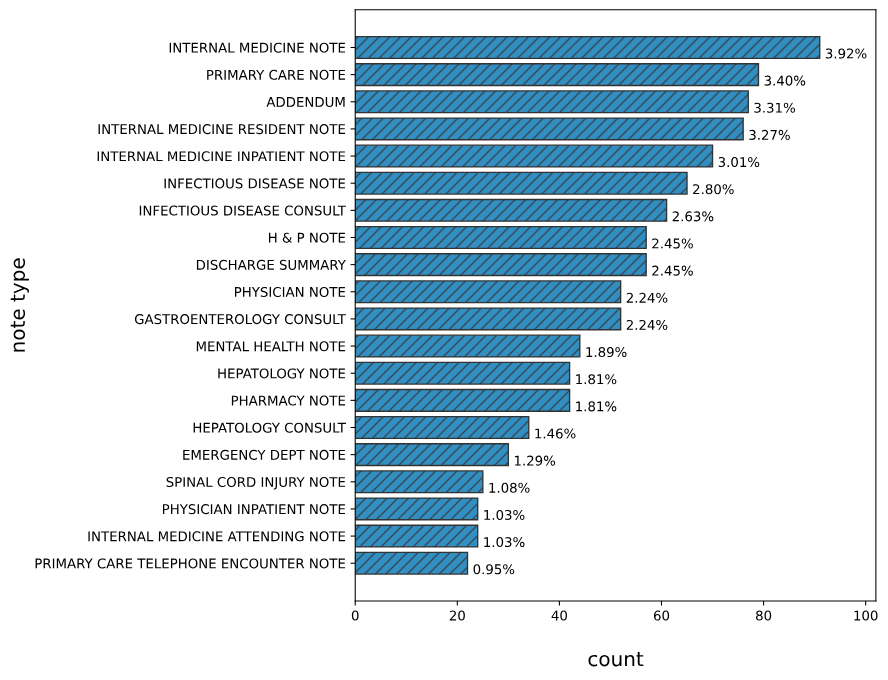}
    \caption{\textcolor{black}{Twenty most frequently encountered clinical note types in this study, along with their frequency distribution.}}
    \label{fig:note_type}
\end{figure}

\textcolor{black}{We also analyze the types of notes included in this study.
Our analysis reveals that there are 411 different types of notes.
Figure \ref{fig:note_type} displays 20 most frequently encountered note types in this study.
Notably, internal medicine notes and primary care notes emerge as the two most prevalent types.
We also find that addendum notes rank third in frequency.
Addendum notes serve as supplements to notes of other types.
}

\begin{figure}[htbp]
    \centering
    \includegraphics[width=.8\textwidth]{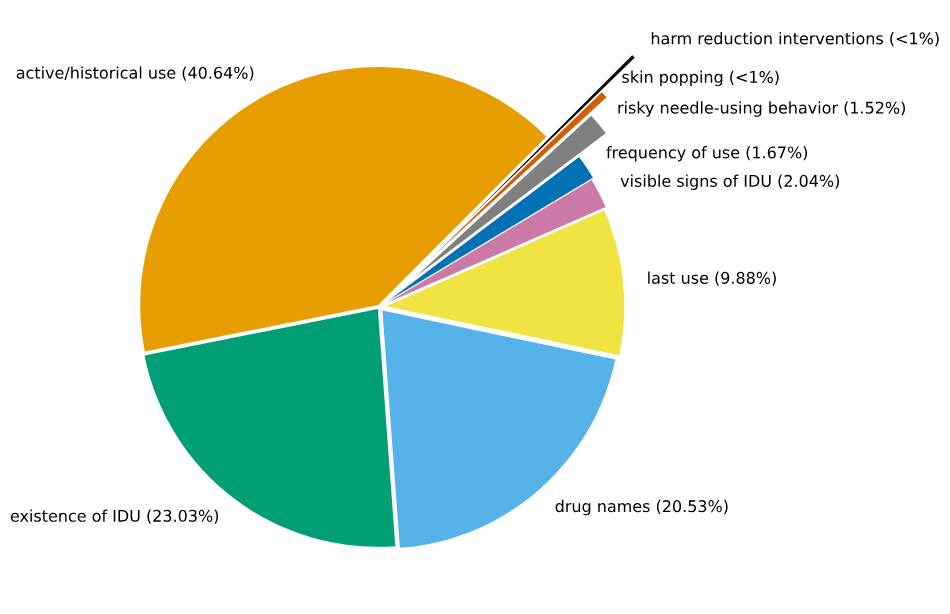}
    \caption{\textcolor{black}{Distribution of the query groups in the gold-standard dataset}}
    \label{fig:pie_cat}
\end{figure}

Table~\ref{tab:data_stat} shows the statistics of our gold-standard dataset.
Our cohort consists of \textcolor{black}{1145} patients with a total of \textcolor{black}{2323} notes that have an average length of \textcolor{black}{1013} words.
\textcolor{black}{Words are identified based on whitespace.}
In addition, we examine the distribution of the query groups outlined in Tables \ref{tab:question} and \ref{tab:qa_map} within the gold-standard dataset.
This analysis is illustrated by the pie chart depicted in Figure \ref{fig:pie_cat}.
The dataset is dominated by QA pairs related to the ``\textcolor{black}{active/historical use}'', as demonstrated.
Following closely behind are QA pairs about \textcolor{black}{``existence of IDU'' and} ``drug names'', whereas the least frequent QA pairs in the dataset are those pertaining to ``skin popping'' and ``harm reduction interventions''.


\begin{table}[htb]
\centering
\begin{tabular}{p{0.35\linewidth}  p{0.5\linewidth}} \hline
\textbf{Property}                  & \textbf{Statistics}                                              \\ \hline
\#Patients                         & 1145                                                             \\
\#Notes                            & 2323                                                             \\
\#Notes per patient (average)      & 2.03                                                             \\
\#Samples in dataset               & 17410                                                            \\
\#QA per note (average)            & 7.49                                                             \\
Note length (in words)             & 1013.09 (average), 1029 (median), 1785 (max)                     \\
Question length (in words)         & 6.72 (average), 7 (median), 14 (max)                             \\
Answer length (in words)           & 7.52 (average), 6 (median), 64 (max)                             \\ \hline
\end{tabular}
\caption{\textcolor{black}{Statistics of the gold-standard dataset}}
\label{tab:data_stat}
\end{table}

\subsection*{Experimental setup}
\label{sec:exp_setup}

For experimentation, we divide our gold-standard dataset into train, validation, and test sets using a \textcolor{black}{70-10-20} split based on patients to avoid any data leakage.
To implement the QA models, we use PyTorch \cite{paszke2019pytorch}.
We use the pre-trained language models from the huggingface API \cite{wolf2019huggingface}.

Based on the statistics of our gold-standard dataset, we choose 512 as the maximum sequence length, 20 as the query length, and 100 as the answer length.
After reviewing the hyperparameters utilized in various QA tasks as outlined in \cite{yasunaga2022linkbert, gu2021domain, raj2021bioelectra, lee2020biobert, beltagy-etal-2019-scibert, alsentzer2019publicly, liu2019roberta, peng2019transfer, devlin2018bert, mahbub2023cpgqa}, we set the document stride to 128 and opted for a batch size of 32, a learning rate of $3e^{-5}$ and a training epoch count of 5 for the training configurations.
We performed all experiments using a single GPU on a Linux virtual machine with two GRID V100-32C GPUs.

\subsection*{Metrics}
\label{sec:met}

\textcolor{black}{To assess the performance of QA models in extracting IDU-related information, we utilize strict matching criteria to compute the F1 score \cite{uzzaman2013semeval}.
It involves verifying if the prediction precisely matches the gold-standard answer character by character, resulting in a strict F1 score per sample that can be either 1 or 0.
Additionally, we use a relaxed matching criteria to measure the F1, precision, and recall scores \cite{uzzaman2013semeval}.
A relaxed match determines if there is any overlap between the prediction and the gold-standard answer.
The recall or sensitivity score per sample reveals the proportion of words in the gold-standard answer that is identified correctly in the predicted answer.
Precision or positive predictive value (PPV) score per sample informs us about the proportion of words in the predicted answer that are actually correct.
In the context of QA problem, when calculating these metrics, true positive refers to the count of tokens that both the predicted answer and the gold-standard answer share, false positive represents the number of tokens found solely in the predicted answer, and false negative indicates the number of tokens only in the gold-standard answer and not in the predicted one.
The relaxed F1, precision, or recall scores per sample can range from 0 to 1.
Following \cite{rajpurkar-etal-2016-squad}, we report the macro-averaged F1 score, accompanied by macro-averaged precision and recall scores on the test sets.
}

\section*{Results and Discussion}
\label{sec:resdisc}

\textcolor{black}{In this section, we report and discuss} the findings from the experiments with QA models. 
Furthermore, we conduct a comprehensive error analysis to demonstrate the capabilities and limitations of the QA models in extracting information related to IDU from clinical notes.

\subsection*{Results on gold-standard test set}
\label{sec:exp_res}

This section focuses on examining the experimental outcomes of the QA models and demonstrates their performance on the test set of our gold-standard dataset.
\textcolor{black}{As shown in Table~\ref{tab:baseline}, ClinicalBERT outperforms other BERT-based QA models.
A strict F1 score of 52\% for ClinicalBERT implies that the QA model can extract IDU-related information 52\% of the time with a strict match to the gold-standard answers.
A relaxed recall score of 79\% on the test set suggests that overall there is a substantial degree of word overlap between the predicted answers and gold-standard answers.
We further analyze the recall score in Section \hyperref[sec:analysis_recall]{Analysis of recall score}.
On the other hand, a relaxed precision score of 85\% in the test set indicates that a higher percentage of terms retrieved as answers by the QA model are included in the gold-standard answers.
A relaxed F1 score of 78\% indicates that the ClinicalBERT model can extract a high percentage of correct information while achieving high precision in those extracted answers.}


\begin{table}[!htb]
\centering
{\begin{tabular}{@{\extracolsep{4pt}}l cccc@{}}\hline
\multirow{2}{*}{Model} & Strict Match   & \multicolumn{3}{c}{Relaxed Match}                            \\
\cline{2-2} \cline{3-5} 
                        & F1             & F1             & Precision      & Recall \\ \hline
BERT                    & 48.10          & 75.88          & 81.57          & 78.82                      \\
BioBERT                 & 46.49          & 74.99          & 81.09          & 76.74                      \\
BlueBERT                & 43.03          & 71.21          & 79.14          & 73.77                      \\
ClinicalBERT            & \textbf{51.65} & \textbf{78.03} & \textbf{85.38} & \textbf{79.02}             \\ \hline
\end{tabular}}
\caption{\textcolor{black}{Performance scores of QA models on the test set}}
\label{tab:baseline}
\end{table}

\subsection*{Temporal out-of-distribution testing}

\begin{figure}[htb]
\includegraphics[width=\linewidth]{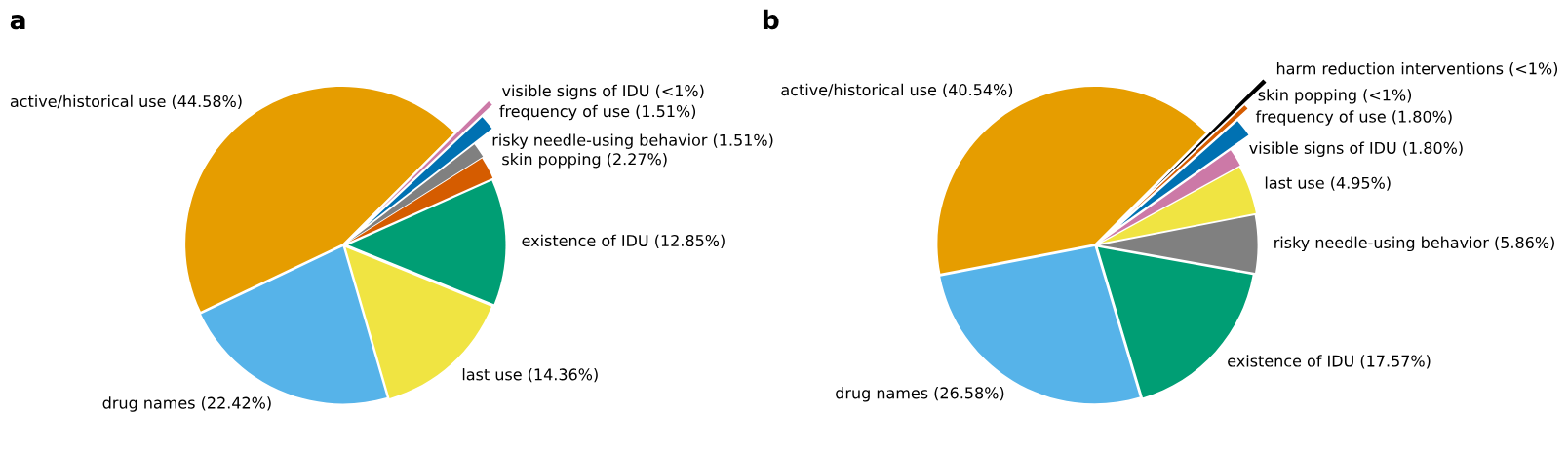}
\caption{Distribution of query groups in additional test datasets built using Cohort-Short (\textbf{a}) and Cohort-Long (\textbf{b}).}
\label{fig:pie_cat_add}
\end{figure}

The writing style of clinical notes may change over time because of changes in clinicians, health care facilities, patients, etc. \cite{gong2022characterizing}.
Given the purpose of our QA model, it is imperative to examine whether the performance of our QA models is retained over time.
Therefore, we perform additional testing of the models from Table~\ref{tab:baseline} on unseen data.
We examine the QA models' short-term and long-term information extraction capabilities by testing on clinical notes from two additional cohorts.
For testing the short-term capability, we \textcolor{black}{randomly select 100 patients and use their} notes from February 2022. \textcolor{black}{Similarly, for} testing the longer-term capability, we \textcolor{black}{randomly select 100 patients and use their} notes from November 2022.
\textcolor{black}{Due to the limitations in our data availability at the time of this study, we were unable to include clinical notes beyond November 2022 for testing the longer-term information extraction capability of the QA models.
In future endeavors, we aim to assess the performance of QA models on more recent notes as part of ongoing research.
}
To avoid data leakage, we use patients and their notes that did not appear in the gold-standard dataset generated by using notes \textcolor{black}{from} January \textcolor{black}{2022}.
We use the method described in Section ~\hyperref[sec:method_dataset]{Gold-standard dataset generation} for building the test datasets using these notes.
\textcolor{black}{Similar to the gold-standard dataset, we manually review these test datasets in collaboration with a subject-matter expert.}
For the rest of the paper, we use the terms ``Cohort-Short'' and ``Cohort-Long'' to represent temporally out-of-distribution notes in February and November, respectively.
Table~\ref{tab:data_stat_add} shows the statistics of the test datasets built using Cohort-Short and Cohort-Long.  
We also show the distribution of query groups in these test datasets in Figure~\ref{fig:pie_cat_add}.
As shown, the distribution of the query groups is similar for the additional test sets and our original gold-standard dataset (refer to Figure~\ref{fig:pie_cat}).


\begin{table}[htb]
\begin{tabular}{p{0.3\linewidth}  p{0.34\linewidth}  p{0.33\linewidth}}
\hline
\multirow{2}{*}{\textbf{Property}}       & \multicolumn{2}{l}{\textbf{Statistics}}                                                         \\ \cline{2-3} 
                                & \textbf{Test Dataset (Cohort-Short)}
                                & \textbf{Test Dataset (Cohort-Long)}                        \\ \hline
\#Patients                                           & 100                      & 100                                                                    \\
\#Notes                                              & 203                      & 146                                                                    \\
\#Notes per patient (average)                        & 2.03                     & 1.46                                                                   \\
\#Samples in dataset                                 & 1985                     & 1110                                                                   \\
\#QA per note (average)                              & 9.78                     & 7.60                                                                   \\
Note length (in words)                               & 1336.53 (average),\newline 1117 (median), 5154 (max) & 1618.86 (average),\newline 1224 (median), 5747 (max)       \\
Question length (in words)                           & 6.51 (average),\newline 7 (median), 14 (max)         & 6.92 (average),\newline 7 (median), 14 (max)               \\
Answer length (in words)                             & 7.59 (average),\newline 7 (median), 21 (max)         & 8.56 (average),\newline 7 (median), 43 (max)      \\ \hline
\end{tabular}
\caption{\textcolor{black}{Statistics of the additional test datasets built using Cohort-Short and Cohort-Long}}
\label{tab:data_stat_add}
\end{table}

Table~\ref{tab:test_feb_nov} shows the performance of the \textcolor{black}{QA} models.
As shown, for both test datasets, \textcolor{black}{the ClinicalBERT model} performs with \textcolor{black}{overall} high scores, reflecting \textcolor{black}{its} competence in extracting information over time.

\begin{table}[htb]
\centering
{\begin{tabular}{@{\extracolsep{4pt}}l cccc@{}}\hline
\multirow{2}{*}{Model} & \multicolumn{4}{c}{Test dataset (Cohort-Short)}                               \\ \cline{2-5} 
                        & Strict Match   & \multicolumn{3}{c}{Relaxed Match}                            \\ \cline{2-2} \cline{3-5} 

                        & F1             & F1             & Precision      & \multicolumn{1}{l}{Recall} \\ \hline
BERT                    & 47.10          & 74.14          & 77.32          & 76.97                      \\
BioBERT                 & 46.95          & 74.25          & 77.36          & \textbf{79.12}             \\
BlueBERT                & 41.76          & 71.71          & 76.47          & 76.18                      \\
ClinicalBERT            & \textbf{55.31} & \textbf{76.12} & \textbf{81.34} & 76.63                      \\ \midrule \midrule
\multirow{2}{*}{Model} & \multicolumn{4}{c}{Test dataset (Cohort-Long)}                               \\ \cline{2-5} 
                        & Strict Match   & \multicolumn{3}{c}{Relaxed Match}                            \\ \cline{2-2} \cline{3-5} 

                        & F1             & F1             & Precision      & \multicolumn{1}{l}{Recall} \\ \hline
BERT                    & 48.38          & 74.38          & 79.57          & 78.99                      \\
BioBERT                 & 50.63          & 73.89          & \textbf{84.44} & 74.15                      \\
BlueBERT                & 51.53          & 73.46          & 81.17          & 78.52                      \\
ClinicalBERT            & \textbf{53.42} & \textbf{75.16} & 81.32          & \textbf{80.20}             \\ \hline
\end{tabular}}
\caption{\textcolor{black}{Performance scores of the QA models on the additional test datasets built using Cohort-Short and Cohort-Long}}
\label{tab:test_feb_nov}
\end{table} 

\subsection*{Error analysis}
\label{sec:error_analysis}
In this section, we provide a comprehensive analysis of the strengths and weaknesses of our best-performing model, which is the \textcolor{black}{ClinicalBERT} QA model, in extracting IDU-related information.
We perform a fivefold analysis as follows: 
Examine the 
(i) confidence intervals of the performance scores,
the effect of  
(ii) note length, (iii) question length, and (iv) gold-standard answer length on the performance of the QA model,
and 
(v) the performance of the QA model for each query group.
Furthermore, \textcolor{black}{analyzing the recall scores,} we showcase the proficiency of the QA model in retrieving IDU-related information.
For our error analysis, we consider all three of our test sets---the test set in our gold-standard dataset and the test datasets from Cohort-Short and Cohort-Long.

\paragraph{Confidence Intervals of Performance Scores:}
We calculate the confidence intervals (CI) for \textcolor{black}{strict F1 score and relaxed F1, precision, and recall} scores achieved by the best-performing QA model to represent how ``good'' these estimates are and thus quantify their uncertainty.
Smaller confidence intervals demonstrated in Table~\ref{tab:conf_int} indicate that our estimates are precise with a high level (95\%) of confidence.

\begin{table}[!htb]
\centering
{\begin{tabular}{@{\extracolsep{4pt}}l cccc@{}}\hline
\multirow{2}{*}{Model}      & Strict Match           & \multicolumn{3}{c}{Relaxed Match}                                                      \\ \cline{2-2} \cline{3-5}
                            & F1 (95\% CI)           & F1 (95\% CI)           & Precision (95\% CI)    & \multicolumn{1}{c}{Recall (95\% CI)} \\ \hline
Gold-standard    & 51.65 (49.92--53.39) & 78.03 (76.99--79.08) & 85.38 (84.40--86.37) & 79.02 (77.89--80.15)               \\
Cohort-Short & 55.31 (53.10--57.10) & 76.12 (74.75--77.48) & 81.34 (80.01--82.67) & 76.63 (75.19--78.06)               \\
Cohort-Long  & 53.42 (50.45--56.35) & 75.16 (73.29--77.03) & 81.32 (79.50--83.14) & 80.20 (78.28--82.12)               \\ \hline
\end{tabular}}
\caption{\textcolor{black}{Performance scores (with 95\% confidence intervals) of the best-performing ClinicalBERT QA model on the test set in the gold-standard dataset and test datasets from Cohort-Short and Cohort-Long}}
\label{tab:conf_int}
\end{table}

\begin{figure}
    \centering
    \includegraphics[width=\textwidth]{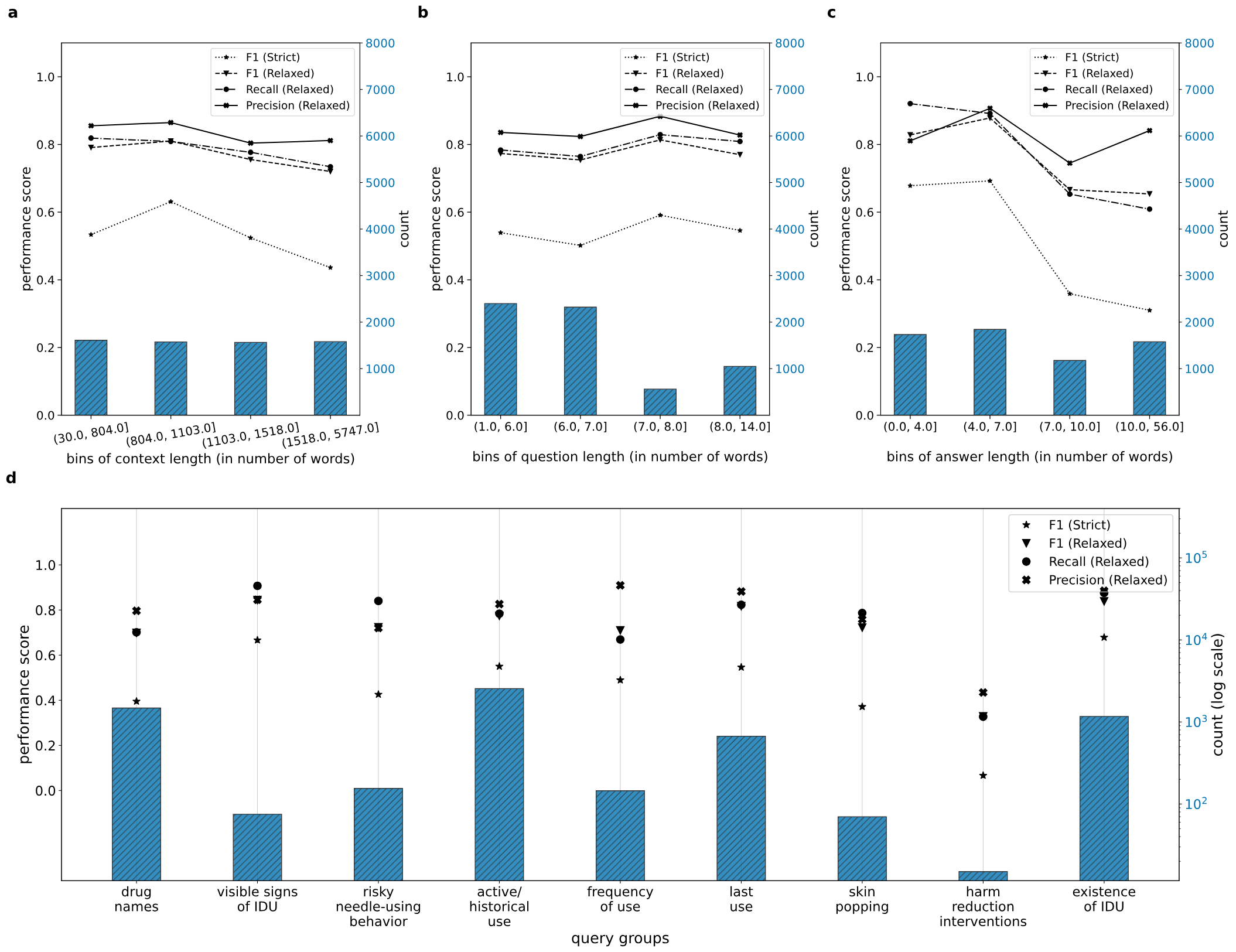}
    \caption{Error analysis. Effect of
    context (note) length (\textbf{a}),
    question length (\textbf{b}), and
    gold-standard answer length (\textbf{c}) on QA model's performance.
    \textbf{d} Performance of QA model for each query group.
    }
    \label{fig:error}
\end{figure}

\paragraph{Effect of Note Length:}
Clinical notes have varying lengths---they can be as short as \textcolor{black}{30} words up to as long as \textcolor{black}{5747} words, based on the statistics of our test datasets.
Therefore, we want the QA model to perform consistently well for all lengths of clinical notes.
To identify the effect of note length on the QA model's performance, we calculate the length of the contexts (i.e., notes) in the three test sets and bin them into four quartiles based on their ascending lengths.
The x-axis in Figure~\ref{fig:error}a shows the length range of these bins, whereas the green bars with the right y-axis show the sample count for each bin.
We find that note length does not have any notable effect on the model's performance scores, depicted on the left y-axis of Figure~\ref{fig:error}a.

\paragraph{Effect of Question Length:}
We also examine the effect of question length on the performance of the QA model.
For this analysis, we adopt the same binning approach as the one on note length.
Figure~\ref{fig:error}b shows that similar to note length, question length also has no effect on the model's performance scores.

\paragraph{Effect of Gold-standard Answer Length:}
In our test sets, we have varying lengths for the gold-standard answers (i.e., extracted information).
For successful implementation, it is essential for the QA model to be able to extract different lengths of information from the clinical notes.
Using the binning approach described earlier for the analysis on note length, we find that the QA model struggles \textcolor{black}{to extract longer gold-standard answers with a strict match}---demonstrated \textcolor{black}{by the strict F1 score} in Figure~\ref{fig:error}c.
\textcolor{black}{Nevertheless, higher relaxed metric scores demonstrated by the QA model indicate its capability to identify the location of the correct answers.
To improve the QA model's proficiency in extracting longer answers with a strict match, additional research is required.}

\paragraph{Performance for Query Groups:}
Based on the information we are interested in extracting from the clinical notes, we create nine query groups as shown in Table~\ref{tab:question}.
The green bars along with the right y-axis in Figure~\ref{fig:error}d show the sample count \textcolor{black}{(in log scale)} for the query groups in the test sets.
The query group ``\textcolor{black}{active/historical use}'' dominates the datasets, followed by the query group \textcolor{black}{``existence of IDU'' and} ``drug names''.
\textcolor{black}{Interestingly we find that the model performs the best on the query group ``visible signs of IDU'' and ``existence of IDU''.
Presumably, the query group ``visible signs of IDU'' has an overall higher performance despite having the third lowest sample count in the test sets and fifth lowest sample count in the gold-standard dataset, because the information queried by this query group usually have some consistent terms in them such as ``track marks'' or ``needle track marks'' along with some other limited relevant information, for example, ``fresh track marks on his forearms''.
We hypothesize that the information extracted by this query group may be easier for the QA model to comprehend.
However, further evaluation of the QA model is necessary to corroborate this hypothesis.
Figure \ref{fig:error}d also shows that the QA model struggles the most with the group ``harm reduction interventions''.
It may happen because ``harm reduction interventions'' has the least number of samples in the gold-standard dataset, possibly causing difficulty for the model to learn from training samples.
It also has the least number of samples in the test sets to obtain a comprehensive overview of the model's performance.
}

\begin{figure}[htb]
    \centering
    \includegraphics[width=.6\textwidth]{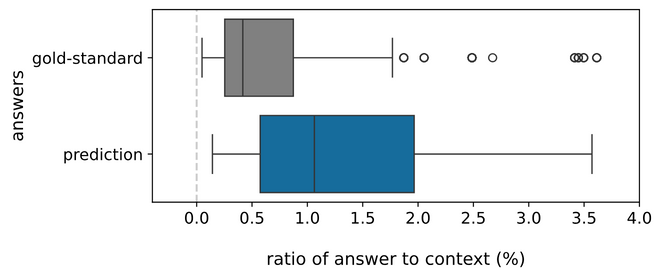}
    \caption{\textcolor{black}{Distributions of the ratios of the predicted answers to the contexts and the ratios of the gold-standard answers to the contexts. The ratios here are presented in the form of percentages.}}
    \label{fig:ratio}
\end{figure}

\paragraph{Analysis of recall score}:
\label{sec:analysis_recall}
\textcolor{black}{In this part of the discussion, we analyze the recall scores of the QA model to shed light on its overall capability to extract gold-standard answers.
In cases where the strict F1 score for the predicted answer is 0, the recall score can demonstrate the overlap between the gold-standard and predicted answers.
For the test set in our gold-standard dataset, our QA model achieved a strict F1 score of approximately 52\%.
For the remaining 48\%, we examine the recall scores by binning them into 12 intervals (shown in Table \ref{tab:recall_analysis}).
We also perform similar analyses for Cohort-Short and Cohort-Long.
As indicated in Table \ref{tab:recall_analysis}, 14\% of the predictions for the gold-standard test set, although lacking a strict match, exhibit a 100
Similarly, for Cohort-Short and Cohort-Long, respectively, 7\% and 15\% of the predicted answers have a 100\% overlap with the gold-standard answers while not having a strict match.
One potential issue while considering 100\% overlap without a strict match is the predicted answer being the entire context. 
To address this concern, we compare the ratio of the predicted answers (that do not have a strict F1 score of 1) to the contexts with the ratio of the gold-standard answers to the contexts.
Figure~\ref{fig:ratio} shows that the distribution of the percentage ratios of the predicted answers to the contexts is similar to that of the gold-standard answers to the contexts.
}

\begin{table}[!htb]
\centering
{\begin{tabular}{@{\extracolsep{4pt}}l cccc@{}}\hline
\multirow{2}{*}{Recall (\%)} &
  \multicolumn{3}{c}{Sample Count (\%)} \\ \cline{2-4}
 &
  \multicolumn{1}{c}{Gold-Standard} &
  \multicolumn{1}{c}{Cohort-Short} &
  \multicolumn{1}{c}{Cohort-Long} \\ \hline
(99--100{]} & 13.76 & 6.65  & 15.32 \\
(90--99{]}  & 0.65  & 0     & 0.45  \\
(80--90{]}  & 2.23  & 0.6   & 0.54  \\
(70--80{]}  & 2.81  & 2.62  & 2.43  \\
(60--70{]}  & 1.33  & 3.43  & 2.7   \\
(50--60{]}  & 2.38  & 1.76  & 3.87  \\
(40--50{]}  & 5.87  & 3.58  & 2.61  \\
(30--40{]}  & 6     & 16.07 & 1.08  \\
(20--30{]}  & 3.37  & 2.87  & 7.39  \\
(10--20{]}  & 4.3   & 1.91  & 4.59  \\
(0--10{]}   & 2.07  & 1.56  & 2.61  \\
0           & 3.56  & 3.63  & 2.97  \\ \hline
\end{tabular}}
\caption{\textcolor{black}{Analysis of recall scores for cases where the predicted answers do not have an strict match with the gold-standard answer.}}
\label{tab:recall_analysis}
\end{table}

\paragraph{Examples of Predicted Answers:}
We demonstrate the capability of the QA model by showing some randomly selected examples of the predicted answers along with the questions and gold-standard answers in Table~\ref{tab:ex_pred}.

\begin{table}[htb]
\centering
\begin{tabular}{p{0.28\linewidth} p{0.36\linewidth} p{.36\linewidth}}\hline
\textbf{Question}                                           & \textbf{Gold-standard Answer}                                      & \textbf{Predicted Answer}                                          \\ \hline
Has the pt ever injected drugs? &
  iv drug use: {[}{]}yes {[}x{]}no &
  iv drug use: {[}{]}yes {[}x{]}no \\ \hline
Which injection drugs? &
  iv heroin &
  attempted to overdose with the intent to die via iv heroin use \\ \hline
Does the patient have any physical evidence of IDU? &
  scars and old track marks &
  old track marks noted on mid arm \\ \hline
Does the pt have a h/o needle-sharing or using dirty needles? &
  sharing needles/injectin heroin/ cocain in the past &
  pos sharing needles \\ \hline
Is the patient actively using IV drugs? &
  h/o iv heroin use about 7 years ago one time, but denies any iv drug use since &
  remote h/o iv heroin use about 7 years ago one time, but denies any iv drug use since. \\ \hline
What is the frequency of pt's injection drug use? &
  daily heroin use with extensive ivdu &
  daily heroin use with extensive ivdu \\ \hline
Does the patient have any history of skin popping? &
 ``skin popping'' heroin 3-4 days ago &
  admits to ``skin popping'' heroin 3-4 days ago and showed the undersigned a keloid scar on his left hand due to shooting heroin \\ \hline
When has the pt last injected drugs? &
  quit iv heroin and cocaine x 6 weeks &
  quit iv heroin and cocaine x 6 weeks \\ \hline
Has the patient received any harm reduction interventions specific to IDU behaviors? &
  topic covered: hiv, viral hepatitis, safe sex practices, pre-exposure prophylaxis syringe service program, naloxone usage &
  syringe service program   \\ \hline
\end{tabular}
\caption{\textcolor{black}{Examples of answers predicted by the QA model along with the question and the gold-standard answers.
}}
\label{tab:ex_pred}
\end{table}

\paragraph{Analysis of Model's Capability to Identify Whether a Note Contains IDU-related Information or Not:}
Our study focuses on extracting IDU-related information from clinical notes, but ideally, we also want our QA model to identify whether the note contains IDU-related information or not.
As such, as an additional analysis, we examine the QA model's ability to identify clinical notes that do not contain any mention of IDU keywords (Table~\ref{tab:keywords}) and as such are assumed to have no information about IDU.
We hypothesize that given a clinical note with no mentions of IDU, the QA model should return an empty string because it could not find the information it was asked to retrieve.

To test this, we use patients from the test set in the gold-standard dataset.
Recall that in our context processing step in Section~\hyperref[sec:method_context]{Note enrichment}, we remove notes that do not contain any IDU keywords.
For this analysis, we incorporate \textcolor{black}{443} notes from \textcolor{black}{226} patients with no mentions of IDU keywords.
We ensure that the notes only belong to the patients in the test set.

To annotate these notes, we use the query group ``existence of IDU'' as questions and empty strings as answers.
For example, given a note with no mentions of IDU and the question ``Has the pt ever injected drugs?'', the QA model should return an empty string.

To measure the performance, we consider only the \textcolor{black}{strict F1 score}.
Thus, if the predicted answer matches with the empty string, we consider that a success (\textcolor{black}{strict F1 score} = 1) and otherwise a failure (\textcolor{black}{strict F1 score} = 0).
We find that our QA model can identify \textcolor{black}{approximately 88\%} of the clinical notes that do not contain any IDU-related information.
\textcolor{black}{Interestingly, we find that for 10\% of the mispredicted answers, the model returned the string ``empty''.
Additionally, we observe that the model returned the string with a single period ``.'', constituting the second most frequently mispredicted answer, accounting for 0.5\% of the predictions.}
Therefore, we can say that while our QA model can extract IDU-related information from clinical notes, it also has the potential to identify the notes that do not contain any.

\textcolor{black}{
\section*{Study limitations}
This study has some limitations. First, the QA model was trained and tested on a dataset that had already undergone a fair amount of NLP pre-processing. Therefore, the model's performance may be limited when generalized to raw, source clinical notes. Further evaluation is needed to prove otherwise.
Second, in many cases, we have noticed the use of terms ``patient denied” or ``veteran tells me” for IDU-related information in the clinical notes. The QA model’s capabilities are limited to the text from which it can extract the pertinent information. Therefore, the QA model must be implemented with supervision in the clinical setting.
Third, our list of IDU keywords/phrases provided by SMEs to filter notes for generating gold-standard dataset is not exhaustive. Notably, drug names such as ``fentanyl'' or ``xylazine'' are absent from the list.
Further assessment is required to measure the QA model's capability to extract information related to these substances.
Fourth, the datasets used in this study have been manually reviewed by one reviewer. Including a second reviewer in the manual review process may ensure more diverse perspectives, reducing the likelihood of individual biases or errors.
}

\section*{Conclusion}

Detection of injection drug use (IDU) behavior among patients is crucial for informed patient care.
In this paper, we tackle the challenging task of IDU-related information extraction from clinical notes.
We build a QA system that takes in a clinical note and an end-user query on IDU and returns the information on IDU extracted from the note.
\textcolor{black}{We hope to potentially integrate the QA model from this study into a user-friendly chatbot framework, enabling clinicians to inquire about information related to nine categories, as identified in this study, with a view to collecting IDU evidence through an interactive platform.}
We evaluate our QA system on a gold-standard dataset built using clinical notes from VA CDW and a combination of manual exploration, rule-based NLP techniques, \textcolor{black}{and subject-matter expert validation}.
We also perform an additional evaluation to examine the capability of our QA model to extract information from temporally out-of-distribution notes.
We then investigate the strengths and limitations of the QA model and identify potential avenues for future research by performing rigorous error analysis.

\textcolor{black}{We have identified the following next steps for this research:
(i) Examine the QA model's capability to extract information from temporally out-of-distribution clinical notes by testing the model on a more recent set of clinical notes.
(ii) Examine/enhance the QA model's capability to handle raw clinical notes without the data-cleaning steps.
(iii) Examine/enhance the QA model's capability to extract information on illicit injection drugs that are not covered in this study, for example, xylazine.
(iv) The extractive QA problem may benefit from the named entity recognition (NER) task \cite{nadapana2022investigating, liu2022qaner}. Subsequent research could explore the integration of NER into the QA task for further investigation.
(v) Expand the applications of QA tasks to extract other types of information from clinical notes, such as information related to alcohol use disorder and substance use disorder. 
}
We hope this method can support the accurate and efficient detection of \textcolor{black}{people who inject drugs and relevant information extraction using their} clinical notes.

\section*{Data availability}
The dataset developed for this study is not accessible to the public under requirements of the Health Insurance Portability and Accountability Act of 1996 and related privacy and security concerns. The underlying electronic health record data can only be used towards improving treatment for patients receiving services from the Veterans Health Administration (VHA). Those interested in accessing VHA EHR data extracts curated for this quality improvement project to replicate and validate findings may contact the corresponding author regarding access via VHA collaboration.

\section*{Code availability}
The code used to develop the QA models is a modified version of the publicly available huggingface example for the question-answering task, which can be found here: \url{https://github.com/huggingface/transformers/blob/master/examples/legacy/question-answering/run_squad.py}.
The modified code is stored in a GitHub repository at \url{https://github.com/mmahbub/qa-system-for-injection-drug-use} \cite{maria_mahbub_2023_10428212}.

\section*{Acknowledgements}

\paragraph{Funding:}
This work was supported by Department of Veterans Affairs, Office of Mental Health and Suicide Prevention. This research used resources of the Knowledge Discovery Infrastructure at the Oak Ridge National Laboratory, which is supported by the Office of Science of the U.S. Department of Energy under Contract No. DE-AC05-00OR22725 and the Department of Veterans Affairs Office of Health Informatics and by VA-DoD Joint Incentive fund under IAA No. 36C10B21M0005.
\\
\textit{Disclaimer:} The views and opinions expressed in this manuscript are those of the authors and do not represent those of the Department of Veterans Affairs, the Department of Energy, or the United States Government.
\\
\textit{General acknowledgements:} The authors wish to acknowledge the support of the larger partnership. Most importantly, the authors would like to thank and acknowledge the veterans who chose to get their care at the VA.

\paragraph{Notice:} This manuscript has been authored by UT-Battelle, LLC, under contract DE-AC05-00OR22725 with the US Department of Energy (DOE). The US government retains and the publisher, by accepting the article for publication, acknowledges that the US government retains a nonexclusive, paid-up, irrevocable, worldwide license to publish or reproduce the published form of this manuscript, or allow others to do so, for US government purposes. DOE will provide public access to these results of federally sponsored research in accordance with the DOE Public Access Plan (\url{https://www.energy.gov/doe-public-access-plan}).

\section*{Author contributions}
M.M., I.D., and E.B. conceptualized the study.
M.M. designed the study, developed the study pipeline and software, generated the QA dataset from the clinical notes, performed visualization, and prepared the manuscript with input from all authors.
I.G. curated the raw clinical notes from VA CDW.
M.M. and I.D. performed the formal analysis of the results.
S.T., K.R., and J.T. assisted in QA dataset generation.
H.S. validated the QA dataset.
I.G., I.D., K.K., S.S., S.T., K.R., H.S., S.M., J.T., E.B., and G.P. reviewed the manuscript and provided feedback.
E.B. acquired funding.
E.B. and G.P. supervised the primary author.

\section*{Competing interests}
The authors declare no competing interests.

\bibliographystyle{unsrt}


\end{document}